\documentclass[11pt,a4paper]{article}
\usepackage[hyperref]{acl2018}
\usepackage{times}
\usepackage{latexsym}
\usepackage{amsmath}

\usepackage{url}

\hypersetup{breaklinks=true}

\aclfinalcopy 
\def\aclpaperid{859} 

\title{Learning from Chunk-based Feedback in Neural Machine Translation}

\author{Pavel Petrushkov \and Shahram Khadivi \and Evgeny Matusov \\
  eBay Inc. \\
Kasernenstr. 25 \\
52064 Aachen, Germany \\
  {\tt \{ppetrushkov, skhadivi, ematusov\}@ebay.com} \\
}

\date{}

\begin{document}
\maketitle
\begin{abstract}

We empirically investigate learning from partial feedback in neural machine translation (NMT), when
partial feedback is collected by asking users to highlight a correct chunk of a translation. 
We propose a simple and effective way of utilizing such feedback in NMT training. 
We demonstrate how the common machine translation problem of domain mismatch between training and deployment can be reduced
solely based on chunk-level user feedback. We conduct a series of simulation experiments to test the effectiveness of the proposed method. 
Our results show that chunk-level feedback outperforms sentence based feedback by up to 2.61\% BLEU absolute.
\end{abstract}

\section{Introduction}
In recent years, machine translation (MT) quality improved rapidly, especially because of advances in neural machine translation (NMT). Most of remaining MT errors arguably come from domain, style, or terminology mismatch between the data on which the MT was trained on and data which it has to translate. It is hard to alleviate this mismatch since usually only limited amounts of relevant training data are available. Yet MT systems deployed on-line in e.g. e-commerce websites or social networks can benefit from user feedback for overcoming this mismatch. Whereas MT users are usually not bilingual, they likely have a good command of the target language and are able to spot severe MT errors in a given translated sentence, sometimes with the help of e.g. an accompanying image, video, or simply prior knowledge. 


A common approach to get user feedback for MT is explicit ratings of translations on an n-point Likert scale. 
The main problem of such methods is that users are not qualified enough to provide reliable feedback for the whole sentence. 
Since different users do not adhere to a single set of guidelines, their ratings may be influenced by various factors, such as
user expectations, user knowledge, or user satisfaction with the platform.
In \cite{kreutzer_new}, the authors investigate the 
reliability and validity of real user ratings by re-evaluating five-star ratings by three independent human annotators,
however the inter-annotator agreement between experts was relatively low and no correlation to the averaged user rating was found.


Instead of providing a rating, a user might be asked to correct the generated translation, in a process called post-editing.
Using corrected sentences for training an NMT system brings larger improvements, but this method requires significant effort and expertise from the user.

Alternatively, feedback can be collected by asking users to mark correct parts (chunks) of the translation 
\cite{marie-max:2015:EMNLP}. 
It can be seen as the middle ground between quick sentence level rating and more expensive post-editing. 
We hypothesize that collecting feedback in this form implicitly forces guidelines on the user, making it less 
susceptible to various user-dependent factors. We expect marking of correct chunks in a translation to be simple enough 
for non-experts to do quickly and precisely and also be more intuitive than providing a numerical rating.

In this paper, we investigate the empirical hypothesis that NMT is able to learn from the good chunks of a noisy sentence and 
describe a simple way of utilizing such chunk-level feedback in NMT training. 
To the best of our knowledge, no dataset with human feedback recorded in this form is available, therefore we experiment with user 
feedback that was artificially created from parallel data.

The rest of this paper is structured as follows. In Section \ref{sec:related} we review related work. We describe 
our partial feedback approach in Section \ref{sec:method}. Next we present our experimental results in 
Section \ref{sec:experiments}, followed by the conclusion in Section \ref{sec:conclusion}.

\section{Related work} \label{sec:related}
Integrating user ratings in NMT has been studied in    
\cite{DBLP:conf/acl/KreutzerSR17}, who view this 
as a bandit structured prediction task.
They demonstrate how the user feedback can be integrated into NMT training and perform a series of experiments using 
GLEU \cite{45610} to simulate user feedback.
Nguyen et al.~\shortcite{DBLP:conf/emnlp/NguyenDB17} 
have also studied this problem and 
adapted an actor-critic approach \cite{pmlr-v48-mniha16} which has shown to be 
robust to skewed, high variance feedback from real users.

\cite{1805.01553} extended the work of \cite{DBLP:conf/emnlp/NguyenDB17} by asking users to provide feedback 
for partial hypotheses to iteratively generate 
the translation, their goal is to minimize the required human involvement. They performed simulated experiments using chrF 
\cite{DBLP:conf/wmt/Popovic15} as simulated feedback.

In all previous works feedback needs to be generated on-line during the training process, however in this paper
we focus on the case where 
there might be a significant time lag
between generation of translation
and acquiring of the feedback.
Lawrence et al.~\shortcite{DBLP:conf/emnlp/LawrenceSR17} 
have proposed a method to leverage user feedback that is available only for logged translated data for a phrase-based 
statistical machine translation system.

\cite{kreutzer_new} have experimented with sentence level star ratings collected from real users of an e-commerce site 
for logged translation data, but found the feedback
to be too noisy to gain improvements.
They also proposed using implicit word level task feedback based on query matching
in an e-commerce application to improve both translation quality and task specific metrics.

Marie and Max~\shortcite{marie-max:2015:EMNLP} have proposed an interactive framework which iteratively improves translation
generated by the phrase-based system by asking users to select correct parts. 
Domingo et al.~\shortcite{W16-3415} extended this idea to also include word deletions and substitutions with the goal of reducing human effort
in translation.

Grangier and Auli~\shortcite{DBLP:journals/corr/abs-1711-04805} 
have studied the task of paraphrasing
an already generated translation by excluding words that the user has marked as incorrect.
They modify NMT model to also accept the marked target sentence as input
and train it to produce similar sentences that do not contain marked words.

\cite{W17-3205,D17-1155} have proposed sentence level weighting method for domain adaptation in NMT.
%
%

\section{Method} \label{sec:method}

In this work we use the encoder-decoder NMT architecture with attention, proposed by 
\cite{DBLP:journals/corr/BahdanauCB14,Sutskever:2014:SSL:2969033.2969173}. 
NMT model is trained to maximize the conditional likelihood of a target sentence
$e_1^I: e_1, \dots, e_I$ given a source sentence $f_1^J: f_1, \dots, f_J$ from
a parallel dataset $D$:

\begin{equation} \label{eq:nmt_loss}
\mathcal{L} = \sum_{f_1^J, e_1^I \in D} \sum_{i=1}^I \log p(e_i | e_1^{i-1}, f_1^J) .
\end{equation}
Training objective \eqref{eq:nmt_loss} is appropriate when the target sentence $e_1^I$ comes from
real data. However, we would like to benefit from model-generated sentences $\tilde{e}_1^I$
by introducing partial feedback.

We assume that partial feedback for a sentence $\tilde{e}_1^I$ is given as a sequence of binary values $w_1^I : w_1, \dots, w_I$,
such that
$w_i=1$ if the word $\tilde{e}_i$ is marked as correct, $w_i=0$ if it is unrated or incorrect.
We propose a simple modification to the loss in Equation \eqref{eq:nmt_loss}:

\begin{equation} \label{eq:nmt_loss_pf}
\mathcal{L}_{PF} = \sum_{f_1^J, \tilde{e}_1^I, w_1^I \in D} \sum_{i=1}^I w_i \log p(\tilde{e}_i | \tilde{e}_1^{i-1}, f_1^J)
\end{equation}
Considering the definition of the binary partial feedback, the model would be trained
to predict correct target words, while ignoring unrated and incorrect ones.
However, incorrect words are still used as inputs to the model and influence 
the prediction context of correct words.

While partial feedback is gathered in a binary form (selected/not selected), word weights $w_i$ can take 
any real value, depending on the weight assignment scheme.

Our training objective can be seen as a generalization of sentence level weighting method \cite{W17-3205,D17-1155}.
The special case of sentence level weight can be expressed as $w_i = w, \forall i$ , where $w$ is the weight for
sentence $\tilde{e}_1^I$.


We differentiate between two practical methods of obtaining the partial feedback data.
First, gathering the feedback from humans, by presenting them with translations and
asking to highlight correct words. This method is expected to produce high quality feedback,
but is relatively expensive and, to the best of our knowledge, no such dataset is publicly available.

Another method is to generate partial feedback automatically using heuristics or statistical models.
This type of feedback would be cheap to obtain, but is unlikely to be of high quality.

In this paper, to show the effectiveness of high quality chunk feedback, we generate  
artificial feedback 
by comparing model predictions to reference
translations using heuristic methods.
This approach is cheap, produces high quality feedback,
but is not practically useful, because it requires access to reference human translation.

We have experimented with several methods of extracting artificial feedback. A simple matching method assigns $w_i=1$
if predicted word $\tilde{e}_i$ is present in reference translation at any position, and $w_i=0$ otherwise.
A slightly more sophisticated method is to find the longest common substring (LCS) between the predicted and reference
translations and set the weights for words which belong to the LCS to 1, and to 0 otherwise. In our experiments we have found the latter method to perform slightly better.

\section{Experiments} \label{sec:experiments}

%

In this section, we conduct a series of experiments to study how well an NMT system is able to learn only from partial user feedback when this feedback is given for in-domain translations, whereas the baseline system is trained on out-of-domain data. 


\subsection{Datasets}
We report results on two datasets: WMT 2017 German to English news translation task \cite{bojar-EtAl:2017:WMT1}
and an in-house  English to Spanish dataset in the e-commerce domain.
On all data we apply byte-pair encoding \cite{DBLP:conf/acl/SennrichHB16a} with 40,000 merge operations learned separately for each language.

For each dataset we separate the larger out-of-domain and smaller in-domain training data.
For De-En we use 1.8M sentence pairs randomly sampled
from available parallel corpora as out-of-domain data and 800K sentence pairs sampled from back-translated
monolingual and unused parallel corpora as in-domain data. For En-Es we have 2.7M  out-of-domain and 1.5M in-domain sentence pairs.
We evaluate our models on \textit{newstest2016} (2999 sentence pairs) for the De-En task and an in-house test set of 1000 sentence pairs
for the En-Es task using case-insensitive BLEU \cite{Papineni:2002:BMA:1073083.1073135} and TER \cite{Snover06astudy}.


We have implemented our NMT model using TensorFlow \cite{tensorflow2015-whitepaper} library. Our encoder is a bidirectional LSTM with 
a layer size of 512; our decoder is an LSTM with 2 layers of the same size. We also use embedding size of 512 and MLP attention layer. 
We train our networks using SGD with a learning rate schedule that starts gradually decaying to 0.01 after the initial 4 epochs.
As regularization we use dropout on the RNN inputs with dropping probability of 0.2.

\subsection{Results} \label{sec:results}
We pre-train baseline NMT models on parallel out-of-domain data for 15 epochs.
We then use the pre-trained model to generate translations from the source side of parallel in-domain corpus.
Using heuristics described in Section \ref{sec:method} and the reference target side of the in-domain corpus we generate artificial
partial feedback to simulate real user input. Then we continue training with a small learning rate for another 
10 epochs on in-domain data with or without user feedback.

\begin{table}[t]
\begin{center}
\begin{tabular}{lcc|cc}
\hline 
 & \multicolumn{2}{c}{De-En} & \multicolumn{2}{c}{En-Es} \\ \cline{2-5}
 &  BLEU & TER & BLEU & TER \\ 
 &  [\%] & [\%] & [\%]  & [\%] \\ \hline
 
Baseline & 30.6 & 49.6 & 32.7 & 52.6\\ 
 ~+self-training  & 31.4 & 48.1 & 35.6 & 49.1 \\

 ~+sent-sBLEU   & 31.4  & 48.1 & 36.0 & 48.4 \\
 ~+sent-binary   & 31.6 & 47.8 & 36.2 & 47.6 \\
 ~+chunk-match & 32.2 & 47.0 & 37.9 & 45.4 \\
 ~+chunk-lcs & \bf 32.3 & \bf 46.5 & \bf 38.8 & \bf 44.5 \\


\hline
\end{tabular}
\end{center}
\caption{\label{tab:pf} Chunk-level feedback compared to sentence-level feedback.  \textit{Self-training} is equivalent to having no feedback or setting all $w_i=1, \forall i$
in the training objective in Eq.~\eqref{eq:nmt_loss_pf}. \textit{sent-sBLEU} and \textit{sent-binary} are sentence-level methods with sentence BLEU and binary weighting rules, defined
as in Section \ref{sec:results}. \textit{chunk-match} and \textit{chunk-lcs}-level feedback refers to assigning $w_i$ using simple 
matching or LCS method described in Section \ref{sec:method}. }
\end{table}

In Table \ref{tab:pf}, we show the effect of different types of feedback on translation performance. First, we see that even using no feedback
slightly improves the model due to self-training on automatically translated in-domain data.

Introducing sentence level feedback improves De-En and En-Es models 
by at most 0.2\% and 0.6\% absolute BLEU, respectively. Sentence level feedback is artificially generated from parallel corpora using 
heuristics, similar to the ones described in Section \ref{sec:method}, but $w_i, \forall i$ are set to the same
sentence weight $w$.
For example, we have tried using sentence BLEU (sBLEU) and a binary rule, which outputs
1 if 
more than 33\% of predicted words 
were marked as correct, and 0 otherwise (binary). 
We have also experimented with
other heuristics, 
but did not achieve better results.


Finally, chunk-based feedback approach based on LCS improves on top of sentence level feedback by another 0.7\% and 2.6\% BLEU for De-En and En-Es, respectively.
We also note a significant improvement of 1.3\% and 3.1\% in TER. Chunk-based approach based on simple matching also outperforms
sentence level methods, but not by as much as lcs-based, which suggests that this method benefits more from consecutive segments,
rather than single correct words.

We believe that the success of the partial feedback approach
can be explained by the fact that often a sentence can be split into chunks which can be translated independently of the context. Reinforcement of the correct translation of such a chunk in one training example seems to positively affect translations of such chunks in other, different sentences.
By focusing on the good and masking out erroneous chunks, partial feedback acts as a precise noise reduction method.

We have also trained the models using fine-tuning \cite{Luong2015StanfordNM} on the reference target in-domain data, which further improved translation by 2\% and 3.8\% BLEU on De-En and En-Es
compared to using chunk-based feedback.
We note that by using partial feedback we are able to recover between 30\% and 45\% of improvements that come from in-domain adaptation. 




\subsection{Robustness}

The proposed artificially generated partial feedback is very precise as it 
does not introduce any type of noise in marking of good chunks.
For example, on the En-Es dataset artificial methods mark 40\% of all words as correct.
However, a user might not mark all the correct words in a sentence,
but select only a few.

Furthermore, artificially generated partial feedback does not contain noise, 
given that the reference translation is adequate. 
However, users may make mistakes in selection.
We differentiate two types of errors that a user can make: under selection, when a correct
word was not marked; and incorrect selection, when an incorrect word was marked as correct.

\begin{table}[t]
\begin{center}
\begin{tabular}{l|lcc|cc}
\hline 
   &  & \multicolumn{2}{c|}{De-En} & \multicolumn{2}{c}{En-Es} \\ \cline{3-6}
\# &	& BLEU & TER & BLEU & TER \\ 
   &  & [\%] & [\%] & [\%]  & [\%] \\ \hline

1 & \multicolumn{2}{l}{Chunk-level} & & & \\
  & ~~~feedback    &  32.3  & 46.5 & 38.8 & 44.5 \\ \hline 
 \multicolumn{6}{l}{Under selection ratio:} \\ \hline
 2 & ~25\%  & 32.2 & 47.0 & 38.9 & 45.0 \\ 
 3 & ~50\%  & 31.9 & 47.4 & 38.1 & 45.6 \\ 
 4 & ~75\%  & 31.4 & 47.9 & 36.7 & 46.7 \\ \hline 

 \multicolumn{6}{l}{Incorrect selection ratio:} \\ \hline
 5 & ~10\% & 32.0 & 47.2 & 38.1 & 44.9 \\ 
 6 & ~25\% & 31.5 & 47.9 & 37.2 & 46.9 \\ 
 7 & ~50\% & 30.9 & 48.8 & 35.6 & 50.0 \\ \hline 
 8 &  \#2 + \#5  & 31.6	 & 47.7 & 38.1 & 45.5 \\ 


\hline
\end{tabular}
\end{center}
\caption{\label{tab:robust} Impact of user errors on the translation performance. \textit{Under selection ratio\%} indicates on average what percentage of words in a correct chunk have not been selected in user simulation, but all selected words are correct. \textit{Incorrect selection ratio\%} indicates what percentage of words are incorrectly selected, here the total number of marked words is the same as in chunk-level feedback. In the last row, 10\% of marked words are actually incorrect and the total number of marked words is 25\% less compared to system in row 1.}
\end{table}

To anticipate the impact of these mistakes we experiment with deliberately impairing the feedback in Table \ref{tab:robust}.
We see that randomly dropping 25\% of the selection has very little effect on the model, while dropping 50\% and more decreases 
the translation performance significantly, yet still performing at the same level or better than self-training system.

When selection contains noise, the impact already becomes noticeable at 10\%. 
Increasing the amount of noise up to 25\% decreases the performance by 1.6\% BLEU in En-Es task.
At 50\% noise level, which is similar to random selection, there is no improvement from using feedback at all.
While we expect users to provide mostly clean feedback, this result indicates the necessity of 
cleaning user feedback data
, e.g. by aggregating feedback from multiple users.

We have also experimented with replacing unselected words by random noise and saw only small decrease in
translation performance, which suggests that our approach is able to benefit from 
very poor translations, as long as the selected chunk is correct.




\subsection{Example}
An example where the NMT system with chunk-based feedback yields a better translation in comparison to other systems is the German sentence 
``Die Krise ist vor\"uber.'' (``The crisis is over.``). The German word ``vor\"uber'' is rare and ambiguous, especially after the BPE-based splitting. 
The system with self-training translates the sentence as ``The crisis is above all.'', whereas the system with chunk-based feedback exactly matches 
the reference translation. We have analyzed the feedback training set: in that data, out of nine occurrences of the word ``vor\"uber'' with the 
reference translation ``over'',  the baseline system got it right three times, getting rewards for the chunks ``is over ...'', ``is over'', ``is over .''
\section{Conclusion and future work} \label{sec:conclusion}

In this work, we have proposed a simple way to integrate partial chunk-based feedback into NMT training. We have experimented with artificially created partial feedback
and shown that using partial feedback results in
significant improvements of MT quality in terms of BLEU and TER. We have shown that chunk-level feedback can be used more effectively than sentence-level feedback. We have studied the robustness of our approach and observed that our model is robust against a moderate amount of noise.

We argue that collecting partial feedback by asking users to highlight correct parts of a translation is more intuitive for users than sentence level ratings and leads to less variation and errors.


In the future, we plan to investigate how to integrate negative partial user feedback, as well as automatic feedback generation methods which do not rely on existing parallel data.




\bibliography{acl2018}
\bibliographystyle{acl_natbib}

\appendix

\end{document}